# Fuzziness, Indeterminacy and Soft Sets: Frontiers and Perspectives

**Michael Gr. Voskoglou** [1]

Department of Applied Mathematics, Graduate Technological Educational Institute of Western Greece,
22334 Patras, Greece; voskoglou@teiwest.gr

**\*** Correspondence: mvoskoglou@gmail.com

:

**Abstract:** The present paper comes across the main steps that laid from Zadeh's fuzziness and Atanassov's intuitionistic fuzzy sets to Smarandache's indeterminacy and to Molodstov's soft sets. Two hybrid methods for assessment and decision making respectively under fuzzy conditions are also presented through suitable examples that use soft sets and real intervals as tools. The decision making method improves an earlier method of Maji et al. Further, it is described how the concept of topological space, the most general category of mathematical spaces, can be extended to fuzzy structures and how to generalize the fundamental mathematical concepts of limit, continuity compactness and Hausdorff space within such kind of structures. In particular, fuzzy and soft topological spaces are defined and examples are given to illustrate these generalizations.

## 1. Introduction

*1.1 Multi-valued Logics*

The development of human science and civilization owes a lot to Aristotle's (384-322 BC) *bivalent logic (BL)*, which was in the center of human reasoning for more than 2 thousand years. BL is based on the "Principle of the Excluded Middle", according to which each proposition is either true or false.

Opposite views were appeared also early in the human history, however, supporting the existence of a third area between true and false, where these two notions can exist together; e.g. by Buddha Siddhartha Gautama (India, around 500 BC), by Plato (427-377 BC), more recently by the philosophers Hegel, Marx, Engels, etc. Integrated propositions of multi-valued logics appeared, however, only during the early 1900s by Lukasiewicz, Tarski and others. According to the "Principle of Valence", formulated by Lukasiewicz, propositions are not only either true or false, but they can have an intermediate truth-value.

*1.2 Literature Review*

Zadeh introduced in 1965 the concept of *fuzzy set (FS)* [1] and developed with the help of it the infinite in the unit interval [0, 1] *fuzzy logic* [2] on the purpose of dealing with partial truths. FL, where truth values are modelled by numbers in the unit interval, satisfies the Lukasiewicz's "Principle of Valence". It was only in a second moment that FS theory and FL were used to embrace *uncertainty* modelling [3, 4]. This happened when membership functions were reinterpreted as possibility distributions. *Possibility theory* is an uncertainty theory devoted to the handling of incomplete information [5]. Zadeh articulated the relationship between possibility and probability, noticing that what is probable must preliminarily be possible [3].

The uncertainty that exists in everyday life and science is connected to inadequate information about the corresponding case. A reduction, therefore, of the existing uncertainty (by a new evidence) means the addition of an equal piece of information. This is why the methods of measuring information (Hartley's formula, Shannon's entropy, etc.) are also used for measuring uncertainty and vice versa; e.g. see [6, Chapter 5].

*Probability* theory used to be for a long period the unique way for dealing with problems connected to uncertainty. Probability, however, is suitable only for tackling the cases of uncertainty which are due to *randomness* [7]. Randomness characterizes events with known outcomes which, however, cannot be predicted in advance, e.g. the games of chance. Starting from the Zadeh's FS, however, various generalizations of FSs and other related theories have been proposed enabling, among others, a more effective management of all types of the existing uncertainty. These generalizations and theories include *type-n FS*, n≥2 [8], *interval-valued FS* [9], *intuitionistic FS (IFS)* [10], *hesitant FS* [11], *Pythagorean FS* [12], *neutrosophic set* [13], *complex FS* [14], *grey system* [15], *rough set* [16], *soft set (SS)* [17], *picture FS* [18], etc. A brief description of all the previous generalizations and theories, the catalogue of which does not end here, can be found in [19].

Fuzzy mathematics have found many and important practical applications (e.g. see [6], [20-24], etc.), but also interesting connections with branches of pure mathematics, like Algebra, Geometry, Topology, etc. (e.g. see [25, 26], etc.).

*1.3 Organization of the Paper*

The paper at hand reviews the process that laid from Zadeh's fuzziness and Atanassov's IFS to Smarandache's indeterminacy and to Molodstov's soft set. It presents also through suitable examples two hybrid methods for *assessment* and *decision making (DM)* under fuzzy conditions using SS and real intervals as tools and describes how one can extend in a natural way the fundamental notion of *topological space (TS)* to fuzzy structures and can generalize the fundamental mathematical concepts of limit, continuity compactness, etc. within such kind of structures. More explicitly, Section 2 contains the basics about FSs and FL needed for this work. In Section 3 the concepts of IFS and NS are defined. The concept of SS is presented in Section 4, where basic operations on SSs are also defined. The hybrid assessment and DM methods are developed in Section 5 and the notion of TS is extended to fuzzy structures in Section 6. The last Section 7 contains the article's final conclusion and some suggestions for future research.

**2. Fuzzy Sets and Fuzzy Logic**

This section contains the basic information about FSs and FL needed for the understanding of the rest of the paper.

*2.1 Fuzzy Sets and Systems*

Zadeh defined the concept of FS as follows [1]:

**Definition 1:** Let U be the universe, then a FS F in U is of the form

$$F = \{(x, m(x)): x \in U\} \qquad (1)$$

In (1) m: U $\to$ [0,1] is the *membership function* of F and m(x) is called the *membership degree* of x in F. The closer m(x) to 1, the better x satisfies the property of F. A crisp subset F of U is a FS in U with membership function such that m(x)=1 if x belongs to F and 0 otherwise.

FSs tackle successfully the uncertainty due to *vagueness*, which is created when one is unable to distinguish between two properties, such as "a good player" and "a mediocre player". A serious disadvantage of FSs, however, is that there is not any exact rule for defining properly their membership function. The methods used for this are usually statistical or intuitive/empirical. Moreover, the definition of the membership function is not unique depending on the "signals" that each person receives from the environment. For example, defining the FS of "old people" one could consider as old all those aged more than 50 years and another one all those aged more than 60 years. As a result the first person will assign membership degree 1 to all people aged between 50 and 60 years, whereas the second will assign membership degrees less than 1. Analogous differences will appear, therefore, to the membership degrees of all the other people. Consequently, the only restriction for the definition of the membership function is that it must be compatible to the common sense; otherwise the resulting FS does not give a creditable description of the corresponding real case. This could happen, for instance, if in the previous example people aged less than 20 years possessed membership degrees $\geq$0.5.

**Definition 2:** The *universal FS* $F_U$ and the *empty FS* $F_\emptyset$ in the universe U are defined as the FSs on U with membership functions m(x)=1 and m(x)=0 respectively, for all x in U.

**Definition 3:** If K and L are FSs in U with membership functions $m_K$ and $m_L$ respectively, then K is called a *fuzzy subset* of L if $m_K(x) \leq m_L(x)$, for all x in U. We write then $K \subseteq L$. If $m_K(x) < m_L(x)$, for all x in U, then K is said to be a *proper fuzzy subset* of L and we write $K \subset L$.

**Definition 4:** If K and L are FSs in U with membership functions $m_K$ and $m_L$ respectively, then:
- The *union* K∪L is said to be the FS in U with membership function $m_{K \cup L}(x)$=max {$m_K(x)$, $m_L(x)$}, for each x in U.
- The *intersection* K∩L is said to be the FS in U with membership function $m_{K \cap L}(x)$=min {$m_K(x)$, $m_L(x)$}, for each x in U.
- The *complement* of K is the FS K* in U with membership function m*(x) = 1- m(x), for all x in U.

If K and L are crisp subsets of U, then all the previous definitions reduce to the ordinary definitions for crisp sets.

Zadeh realized that FSs correspond to words (adjectives or adverbs) of the natural language [27]; e.g. the word "clever" corresponds to the FS of the clever people, since how clever is everyone is a matter of degree. A synthesis of FSs related to each other is said to be a *fuzzy system*, which mimics the way of human reasoning. For example, a fuzzy system can control the function of an air - condition, can send signals for purchasing shares, etc. [21].

*2.2 Probabilistic Vs Fuzzy Logic - Bayesian Reasoning*

Many of the traditional supporters of the classical BL claimed that, since BL works effectively in science and the computers and explains the phenomena of the real world, except perhaps those that happen in the boundaries, there is no reason to introduce the unstable principles of a multi-valued logic. FL, however, aims exactly at clearing the happenings in the boundaries! Look, for example, at Figure 1 [28] representing the FS T of "tall people". People with height less than 1.50 m possess membership degree 0 in T. The membership degrees are increasing for heights greater than 1.50m, taking the value 1 for heights equal or greater than 1.80 m. Therefore, the "fuzzy part" of the graph - which is represented, for simplicity, in Figure 1 by the straight line segment AC, but its exact form depends upon the definition of the membership function- lies in the area of the rectangle ABCD formed by the OX axis, its parallel through the point E and the two perpendicular to the OX lines at the points A and B.

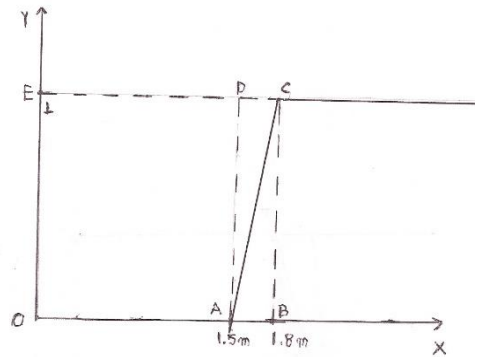

**Figure 1:** Graph of the FS of "tall people"

BL, on the contrary, considers a bound (e.g. 1.8 m) above which people are tall and below which are short. Thus, a individual with height 1.805 m is considered to be tall, whereas another with height 1.795 m is considered to be short!

In conclusion, FL generalizes and completes the traditional BL fitting better not only to our everyday life situations, but also to the scientific way of thinking. More details about FL can be found in Section 2 of [28].

E. Jaynes argued that probability theory is a generalization of BL reducing to it in cases in which something is either absolutely true or absolutely false [29]. A considerable number of scientists, like the famous for his contributions to Algebraic Geometry D. Mumfrord [30], supported his ideas. Nevertheless, as we have already seen in our Introduction, probability, due to its bivalent texture, tackles effectively only the uncertainty due to randomness. Therefore, Jaynes' probabilistic logic is subordinate to FL.

*Bayesian Reasoning*, however, connects BL and FL [31]. In fact, the *Bayes' rule* expressed by equation (2) below, calculates the *conditional probability* P(A/B) with the help of P(B/A), of the *prior probability* P(A) and the *posterior probability* P(B)

$$P(A/B) = \frac{P(B/A)P(A)}{P(B)} \quad (2).$$

The value of P(A) is fixed before the experiment, whereas the value of P(B) is obtained from the experiment's data. Frequently, however, the value of P(A) is not standard. In such cases different values of the conditional probability P(A/B) are obtained for all the possible values of P(A). Consequently, the Bayes' rule tackles the existing, due to the imprecision of the value of the prior probability, uncertainty in a way analogous to FL ([32], Section 5).

Bayesian reasoning is very important in everyday life situations and for the whole science too. Recent researches have shown that most of the mechanisms under which the human brain works are Bayesian [33]. Thus, Bayesian reasoning is a very useful tool for *Artificial Intelligence (AI)*, which mimics the human behavior. The physicist and Nobel prize winner John Mather has already expressed his uneasiness about the possibility that the Bayesian machines could become too smart in future, so that to make humans to look useless [34]! Consequently, Sir Harold Jeffreys (1891-1989) has successfully characterized the Bayesian rule as the "Pythagorean Theorem of Probability Theory" [35].

**3. Intuitionistic Fuzzy Sets and Neutrosophic Sets**

K. Atanassov, Professor of Mathematics at the Bulgarian Academy of Sciences added in 1986 to Zadeh's membership degree the degree of *non-membership* and introduced the concept of IFS as follows [10]:

**Definition 5:** An IFS A in the universe U is defined as the set of the ordered triples

$$A = \{(x, m(x), n(x)): x \in U, 0 \leq m(x) + n(x) \leq 1\} \quad (3)$$

In equation (3) m: $U \to [0,1]$ is the membership function and n: $U \to [0,1]$ is the non-membership function.

We can write $m(x) + n(x) + h(x) = 1$, where $h(x)$ is the *hesitation* or *uncertainty degree* of x. If $h(x) = 0$, then the IFS becomes a FS. The name intuitionistic was given because an IFS has inherent the intuitionistic idea by incorporating the degree of hesitation.

For example, if A is the IFS of the good students of a class and $(x, 0.7, 0.2) \in A$, then x is characterized as a good student by the 70% of the teachers' of the class, as not good by 20% of them, whereas there is a hesitation by 10% of the teachers' to characterize him/her as either good or not good student. Most concepts and operations about FSs can be extended to IFSs, which simulate successfully the existing imprecision in human thinking [36].

F. Smarandache, Professor of the New Mexico University, defined in 1995 the concept of NS as follows [13]:

**Definition 6:** A *single valued NS (SVNS)* A in the universe U has the form

$$A = \{(x, T(x), I(x), F(x)): x \in U, T(x), I(x), F(x) \in [0,1], 0 \leq T(x)+I(x)+F(x) \leq 3\} \quad (4)$$

In equation (4) $T(x)$, $I(x)$, $F(x)$ are the degrees of *truth* (or membership), *indeterminacy* (or *neutrality*) *and falsity* (or non-membership) of x in A respectively, called the *neutrosophic components* of x. For simplicity, we write A<T, I, F>. The word "neutrosophy" is a synthesis of the word "neutral´ and the Greek word "sophia" (wisdom) and means "the knowledge of neutral thought".

For example, let U be the set of the employees of a company and let A be the SVNS of the working hardly employees. Then each employee x is characterized by a neutrosophic triplet (t, i, f) with respect to A, with t, i, f in [0, 1]. For example, $x(0.7, 0.1, 0.4) \in A$ means that the manager of the company is 70% sure that x works hardly, but at the same time he/she has a 10% doubt about it and a 40% belief that x is not working hardly. In particular, $x(0,1,0) \in A$ means that the manager does not know absolutely nothing about x's affiliation with A.

Indeterminacy is defined to be in general everything that exists between the opposites of truth and falsity [37]. In an IFS is $I(x) = 1 - T(x) - F(x)$, i.e. the indeterminacy is equal with the hesitancy. In a FS is $I(x)=0$ and $F(x) = 1 - T(x)$ and in a crisp set is $T(x)=1$ (or 0) and $F(x)= 0$ (or 1). Consequently, crisp sets, FSs and IFSs are special cases of SVNSs.

If T(x) + I(x) + F(x)<1, then it leaves room for incomplete information about x, when is equal to 1 for complete information and when >1 for *paraconsistent* (i.e. contradiction tolerant) information about x. A SVNS may contain simultaneously elements leaving room for all the previous types of information.

If T(x) + I(x) + F(x)<1, $\forall$ x ∈ U, then the corresponding SVNS is called a *picture FS (PiFS)* [18]. In this case 1- T(x)-I(x)-F(x) is the degree of *refusal membership* of x in A. The PiFSs are tackling successfully cases related to human opinions involving answers of types yes, abstain, no and refusal to participate, like the voting processes.

The difference between the *general definition of a NS* and the already given definition of a SVNS is that in the former case T(x), I(x) and F(x) may take values in the non-standard unit interval ]−0, 1+[, which includes values <0 or >1. For example, in a bank with full-time work 35 hours per week one upon his/her work could belong by $\frac{35}{35}$=1 to the bank (full-time) or by $\frac{20}{35}$<1 (part-time) or by $\frac{40}{35}$>1 (over-time). Assume further that an employee caused a damage which is balanced with his salary. Then, if the cost is equal to $\frac{40}{35}$ of his weekly salary, the employee belongs this week to the bank by -$\frac{5}{35}$<0.

Most concepts and operations of FSs and IFSs are extended to NSs [38], which, apart from vagueness, tackle adequately the uncertainty due to *ambiguity* and *inconsistency*. Ambiguity takes place when the available information can be interpreted in several ways. This could happen, for example, among the jurymen of a trial. Inconsistency appears when two or more pieces of information cannot be true at the same time. As a result the obtainable in this case information is conflicted or undetermined. For example, "The probability for being windy tomorrow is 90%, but this does not mean that the probability of not having strong winds is 10%, because they might be hidden meteorological conditions".

For the same reason as for the membership function of a FS there is a difficulty to define properly the neutrosophic components of the elements of the universe in a NS. The same happens in case of all generalizations of FSs involving membership degrees (e.g. IFSs, etc.). This caused in 1975 the introduction of the *interval-valued FS (IVFS)* defined by a mapping from the universe U to the set of closed intervals in [0, 1] [9]. Other related to FSs theories were also developed, in which the definition of a membership function is either not necessary (grey systems/numbers [15]), or it is overpassed either by using a pair of sets which give the lower and the upper approximation of the original crisp set (rough sets [16]), or by introducing a suitable set of parameters (SSs [17]).

## 4. Soft Sets

*4.1 The Concept of Soft Set*

In 1999 D. Molodstov, Professor of the Russian Academy of Sciences, introduced the notion of *soft set (SS)* as a means for tackling the uncertainty in terms of a suitable set of parameters in the following way [17]:

**Definition 7:** Let E be a set of parameters, let A be a subset of E, and let f be a map from A into the power set P(U) of the universe U. Then the SS (f, A) in U has the form

$$(f, A) = \{(e, f(e)): e \in A\} \quad (5)$$

In other words, a SS can be considered as a parametrized family of subsets of U. The name "soft" is due to the fact that the form of (f, A) depends on the parameters of A. For each e ∈ A, its image f(e) in P(U) is called the *value set* of e in (f, A), while f is called the *approximation function* of (f, A).

For example, let U= $\{C_1, C_2, C_3\}$ be a set of cars and let E = $\{e_1, e_2, e_3\}$ be the set of the parameters $e_1$=cheap, $e_2$=hybrid (petrol and electric power) and $e_3$= expensive. Let us further assume that $C_1$, $C_2$ are cheap, $C_3$ is expensive and $C_2$, $C_3$ are the hybrid cars. Then, a map f: E → P(U) is defined by $f(e_1)=\{C_1, C_2\}$, $f(e_2)=\{C_2, C_3\}$ and $f(e_3)=\{C_3\}$. Therefore, the SS (f, E) in U is the set of the ordered pairs (f, E) = $\{(e_1, \{C_1, C_2\}), (e_2, \{C_2, C_3\}), (e_3, \{C_3\})\}$. The SS (f, E) can be represented by the graph of Figure 2.

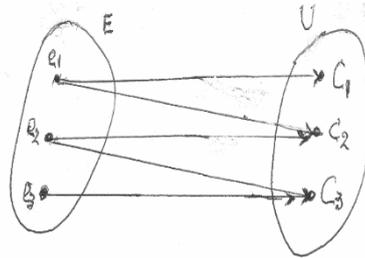

**Figure 2:** Graphical representation of the SS (f, E)

On comparing the graphs of Figures 1 and 2 one can see that a FS is represented by a simple graph, whereas a bipartite graph [39] is needed for the representation of a SS.

Maji et al. [40] introduced a *tabular representation* of SSs in the form of a binary matrix in order to be stored easily in a computer's memory. For example, the tabular representation of the soft set (f, E) is given in Table 1.

A FS in U with membership function y = m(x) is a SS in U of the form (f, [0, 1]), where $f(\alpha)=\{x \in U: m(x) \geq \alpha\}$ is the corresponding *a-cut* of the FS, for each $\alpha$ in [0, 1]. Consequently the concept of SS is a generalization of the concept of FS.

An important advantage of SSs is that, by using the parameters, they pass through the already mentioned difficulty of defining properly membership functions.

**Table 1.** Tabular representation of the SS (f, E)

|  | $e_1$ | $e_2$ | $e_3$ |
|---|---|---|---|
| $C_1$ | 1 | 0 | 0 |
| $C_2$ | 1 | 1 | 0 |
| $C_3$ | 0 | 1 | 1 |

*4.2 Operations on Soft Sets*

**Definition 8:** The *absolute SS* $A_U$ is the SS (f, A) in which f(e)=U, ∀ e ∈ A, and the *null soft set* $A_\emptyset$ is the SS (f, A) in which f(e)=∅, ∀ e ∈A.

**Definition 9:** If (f, A) and (g, B) are SSs in U, (f, A) is a *soft subset* of (g, B), if A$\subseteq$B and f(e)$\subseteq$ g(e), $\forall$ e $\in$A. We write then (f, A)$\subseteq$(g, B). If A$\subset$B, then (f, A) is called a *proper soft subset* of B and we write (f, A)$\subset$(g, B).

**Definition 10:** Let (f, A) and (g, B) be SSs in U. Then:
- The *union* (f, A) $\cup$ (g, B) is the SS (h, A$\cup$B) in U, with h(e)=f(e) if e $\in$ A-B, h(e)=g(e) if e $\in$ B-A and h(e)= f(e)$\cup$g(e) if e $\in$ A$\cap$B.
- The *intersection* (f, A) $\cap$ (g, B) is the soft set (h, A$\cap$B) in U, with h(e)=f(e)$\cap$g(e), $\forall$ e $\in$ A$\cap$B.
- The *complement* (f, A)$^C$ of the soft SS (f, A) in U, is defined to be the SS (f*, A) in U, in which the function f* is defined by f*(e) = U-f(e), $\forall$ e $\in$ A.

For general facts on soft sets we refer to [41].

**Example 1:** Let U={$H_1$, $H_2$, $H_3$}, E={$e_1$, $e_2$, $e_3$} and A= {$e_1$, $e_2$}. Consider the SS
S = (f, A) = {($e_1$, {$H_1$, $H_2$}), ($e_2$, {$H_2$, $H_3$})} of U. Then the soft subsets of S are the following:

$S_1$={($e_1$, {$H_1$})}, $S_2$={($e_1$, {$H_2$})}, $S_3$={($e_1$, {$H_1$, $H_2$})},

$S_4$={($e_2$, {$H_2$})}, $S_5$={($e_2$, {$H_3$})}, $S_6$={($e_2$, {$H_2$, $H_3$})},

$S_7$={($e_1$, {$H_1$}), ($e_2$, {$H_2$})}, $S_8$={($e_1$, {$H_1$}), ($e_2$, {$H_3$})},

$S_9$={($e_1$, {$H_2$}), ($e_2$, {$H_2$})}, $S_{10}$={($e_1$, {$H_2$}), ($e_2$, {$H_3$})},

$S_{11}$={($e_1$, {$H_1$, $H_2$}), ($e_2$, {$H_2$})}, $S_{12}$={($e_1$, {$H_1$, $H_2$}), ($e_2$, {$H_3$})},

$S_{13}$={($e_1$, {$H_1$}), ($e_2$, {$H_2$, $H_3$})}, $S_{14}$={($e_1$, {$H_2$}), ($e_2$, {$H_2$, $H_3$})},

S, $A_\emptyset$={($e_1$, $\emptyset$), ($e_2$, $\emptyset$)}

It is easy also to check that (f, A)$^C$= {($e_1$, {$H_3$}), ($e_2$, {$H_1$})}.

## 5. Hybrid Assessment and Decision Making Methods under Fuzzy Conditions

Each of the various theories that have been proposed for tackling the existing in real world uncertainty [19] is more suitable for certain types of uncertainty. Frequently, however, a combination of two or more of these theories gives better results. To support this argument, we present here two hybrid methods for assessment [42] and decision making [43] respectively under fuzzy conditions using SSs and closed real intervals as tools

*5.1 Using Closed Real Intervals for Handling Approximate Data*

An important perspective of the closed intervals of real numbers is their use for handling approximate data. In fact, a numerical interval I = [x, y], with x, y real numbers, x<y, is actually representing a real number with known range, whose exact value is unknown. When no other information is given about this number, it looks logical to consider as its representative approximation the real value

$$V(I) = \frac{x+y}{2} \quad (6)$$

The closer x to y, the better V(I) approximates the corresponding real number.

Moore et al. introduced in 1995 [44] the basic arithmetic operations on closed real intervals. In particular, and according to the interests of the present article, if $I_1$ = [$x_1$, $y_1$] and $I_2$ = [$x_2$, $y_2$] are closed intervals, then their *sum* $I_1$ + $I_2$ is the closed interval

$$I_1 + I_2 = [x_1+ x_2, y_1+ y_2] \quad (7)$$

Also, if k is a positive number then the *scalar product* $kI_1$ is the closed interval

$$kI_1 = [kx_1, ky_1] \qquad (8)$$

When the closed real intervals are used for handling approximate data, are also referred as *grey numbers (GNs)*. A GN [x, y], however, may also be connected with a *whitenization function*
f: [x, y] → [0, 1], such that, ∀ a ∈ [x, y], the closer f(a) to 1, the better a approximates the unknown number represented by [x, y] [22, Section 6.1].

We close this subsection about closed real intervals with the following definition, which will be used in the assessment method that follows.

**Definition 11:** Let $I_1, I_2,...., I_k$ be a finite number of closed real intervals and assume that $I_i$ appears $n_i$ times in an application, i = 1,2,…., k. Set n = $n_1+n_2+….+n_k$. Then the *mean value* of all these intervals is defined to be the closed real interval

$$I = \frac{1}{n}(n_1I_1+n_2I_2+….+n_kI_k) \qquad (9)$$

*5.2 The Assessment Method*

Assessment is one of the most important components of all human and machine activities, helping to determine possible mistakes and to improve performance with respect to a certain activity. The assessment processes are realized by using either numerical or linguistic (qualitative) grades, like excellent, good, moderate, etc. Traditional assessment methods are applied in the former case which give accurate results, the most standard among them being the calculation of the mean value of the numerical scores.

Frequently, however, the use of numerical scores is either not possible (e.g. in case of approximate data) or not desirable (e.g. when more elasticity is required for the assessment). In such cases assessment methods based on principles of FL are usually applied. A great part of the present author's earlier researches were focused on developing such kind of methods, most of which are reviewed in detail in [22]. It seems, however, that proper combinations of the previous methodologies could give better results (e.g. see [42]).The assessment method developed by the present author in [42] will be illustrated here with the help of the following example

**Example 2**: Let U = $\{p_1, p_2,… , p_{19}, p_{20}\}$ be the set of the players of a football team. Assume that the first 3 of them are excellent players, the next 7 very good players, the following 5 good players, the next 3 mediocre players, and the last 2 new players with no satisfactory performance yet. It is asked: 1) to make a parametric assessment of the team's quality, and 2) to estimate the mean potential of the team.

*Solution:* 1) Consider the linguistic grades A=excellent, B=very good, C=good, D=mediocre, and F=not satisfactory, set E = {A, B, C, D, F} and define a map f: E → P(U) by f(A) = $\{p_1, p_2, p_3\}$, f(B) = $\{p_4, p_5, … ,p_{10}\}$, f(C) = $\{p_{11}, p_{12},… , p_{15}\}$, f(D) = $\{p_{16}, p_{17}, p_{18}\}$, and f(F) = $\{p_{19}, p_{20}\}$. Then the required parametric assessment of the team's quality can be represented by the soft set
(f, E) = {(A, f(A)), (B, f(B)), (C, f(C)), (D, f(D)), (F, f(F))}.

2) Assign to each parameter (linguistic grade) of E a closed real interval, denoted for simplicity by the same letter, as follows: A = [85, 100], B = [75, 84], C = [60, 74], D= [50, 59], F= [0, 49]. Then by (9) the mean potential of the football team can be approximated by the real interval

M = $\frac{1}{20}$ (3A+7B+5C+3D+2F).

Applying equations (7) and (8) and making the corresponding calculations one finds that M= $\frac{1}{20}$ [1230, 1533] = [61.5, 76.65]. Thus, equation (5) gives that V(M) = 69.075, which shows that the mean potential of the football team is good (C).

**Remark 1:** The choice of the intervals in case 2 of the previous example corresponds to generally accepted standards for translating the linguistic grades A, B, C, D, F in the numerical scale 0 -100. By no means, however, this choice could be considered as being unique, since it depends on the special beliefs of the user. For example, one could as well choose A = [80, 100], B = [70, 79], C = [60, 69], D= [50, 59], F= [0, 49], etc.

**Remark 2:** One could equivalently use *triangular fuzzy numbers (TFNs)* instead of closed real intervals in the previous example [22].

*5.3 The Decision Making Method*

Maji et al. [40] developed a parametric DM method using SSs as tools. In an earlier work [42] we have improved their method by adding closed real intervals (GNs) to the tools. Here we illustrate our improved method with the following example.

**Example 3:** A candidate buyer, who believes that the ideal house to buy should be cheap, beautiful, wooden and in the country, has to choose among six houses $H_1$, $H_2$, $H_3$, $H_4$, $H_5$ and $H_6$, which are for sale. Assume further that $H_1$, $H_2$, $H_6$ are the beautiful houses, $H_2$, $H_3$, $H_5$, $H_6$ are in the country, $H_3$, $H_5$ are wooden and $H_4$ is the unique cheap house. Which is the best choice for the candidate buyer?

*Solution:* First we solve this DM problem following the method of Maji et al. [46]. For this, consider U = {$H_1$, $H_2$, $H_3$, $H_4$, $H_5$, $H_6$} as the set of the discourse and let E = {$e_1$, $e_2$, $e_3$, $e_4$} be the set of the parameters $e_1$=beautiful, $e_2$=in the country, $e_3$=wooden and $e_4$=cheap. Then a map f: E → P(U) is defined by f($e_1$) = { $H_1$, $H_2$, $H_6$}, f($e_2$) = { $H_2$, $H_3$, $H_5$, $H_6$}, f($e_3$) = { $H_3$, $H_5$}, f($e_4$) = { $H_4$}, which gives rise to the SS (f, E) = {($e_1$, f($e_1$)), ($e_2$, f($e_2$)), ($e_3$, f($e_3$)), ($e_4$, f($e_4$))}.

One can write the previous SS in its tabular form as it is shown in Table 2.

**Table 2.** Tabular representation of the SS (f, E)

|  | $e_1$ | $e_2$ | $e_3$ | $e_4$ |
|---|---|---|---|---|
| $H_1$ | 1 | 0 | 0 | 0 |
| $H_2$ | 1 | 1 | 0 | 0 |
| $H_3$ | 0 | 1 | 1 | 0 |
| $H_4$ | 0 | 0 | 0 | 1 |
| $H_5$ | 0 | 1 | 1 | 0 |
| $H_6$ | 1 | 1 | 0 | 0 |

Then, the *choice value* of each house is calculated by adding the binary elements of the row of Table 1 in which it belongs. The houses $H_1$ and $H_4$ have, therefore, choice value 1 and all the others have choice value 2. Consequently, the candidate buyer must choose one of the houses $H_2$, $H_3$, $H_5$ or $H_6$.

The previous decision, however, is obviously not so helpful. This gives the hint of revising the previous DM method of Maji et al. In fact, observe that, in contrast to $e_2$ and $e_3$, the parameters $e_1$ and $e_4$ in the present problem have not a bivalent texture. This means that it is closer to reality to characterize them by the qualitative grades A, B, C, D and F of Example 1, than by the binary elements 0, 1.

Assume, therefore, that the candidate buyer, after studying carefully all the existing information about the six houses under sale, decided to use the following Table 3 instead of Table 2 for making the right decision

**Table 3.** Revised tabular representation of the SS (f, E)

|       | $e_1$ | $e_2$ | $e_3$ | $e_4$ |
|-------|-------|-------|-------|-------|
| $H_1$ | A | 0 | 0 | C |
| $H_2$ | A | 1 | 0 | F |
| $H_3$ | C | 1 | 1 | C |
| $H_4$ | D | 0 | 0 | A |
| $H_5$ | D | 1 | 1 | C |
| $H_6$ | A | 1 | 0 | D |

From Table 3 one calculates the choice value $C_i$ of the house $H_i$, i=1, 2, 3, 4, 5, 6 as follows: $C_1$=V(A+C), or by (6) $C_1$= V([0.85+0.6, 1+0.74]) and finally by (5) $C_1 = \frac{1.45+1.74}{2}$ =1.595. Similarly $C_2$=1+V(A+F)=1+ $\frac{0.85+1.49}{2}$ =2.17, $C_3$=2+V(C+C)=3.34, $C_4$=V(D+A)=1.47, $C_5$=2+V(D+C)=3.215, and $C_6$=1+V(A+D)=2.47. The right decision is, therefore, to buy the house $H_3$.

**Remark 3:** One could, as in Example 2, use TFNs instead of closed real intervals [45] in this DM problem.

**Remark 4:** The novelty of our hybrid DM method with respect to the DM method of Maji et al. [40] is that by using closed real intervals instead of the binary elements 0, 1 in the tabular matrix of the corresponding SS in cases where some (or all) of the parameters are not of bivalent texture, we succeed to make a better decision.

*5.4 Weighted Decision Making*

When the goals put by the decision-maker are not of the same importance, weight coefficients must be assigned to each parameter for making the proper decision. Assume, for instance, that in the previous example the candidate buyer assigned the weight coefficients 0.9 to $e_1$, 0.7 to $e_2$, 0.6 to $e_3$ and 0.5 to $e_4$. Then, the weighted choice values of the houses in Example 3 are calculated as follows:

$C_1$=V(0.9A+0.5C), or by (5), (6) and (7) $C_1$=V([1.65, 1.27])=1.46. Similarly $C_2$=0.7+V(0.9A+0.5F) =0.7+V([0.765,1.145])=1.655, $C_3$=0.7+0.6+V(0.9C+0.5C)=1.3+V([0.84,1.036])=2.238, $C_4$=V(0.9D+0.5A) =V([0.875,1.031])=0.953, $C_5$=0.7+0.6+V(0.9D+0.5C)=1.3+V([0.75,0.901])=2.1255, $C_6$=0.7+V(0.9A+0.5D) =0.7+V([1.015, 1.195])=1.805. Consequently, the right decision is again to buy the house $H_3$.

## 6. Topological Spaces in Fuzzy Structures

TSs is the most general category of mathematical spaces, in which fundamental mathematical notions are defined [46]. In this section we describe how the concept of TS is extended to fuzzy structures.

### 6.1 Fuzzy Topological Spaces

**Definition 12** [25]: A *fuzzy topology (FT)* T on a non-empty set U is a family of FSs in U such that:
- The universal and the empty FSs belong to T
- The intersection of any two elements of T and the union of an arbitrary number (finite or infinite) of elements of T belong also to T.

Trivial examples of FTs are the *discrete FT* $\{F_\emptyset, F_U\}$ and the *non-discrete FT* of all FSs in U. Another example is the set of all *constant FSs* in U, i.e. all FSs in U with membership function defined by m(x)=c, for some c in [0, 1] and all x in U.

The elements of a FT T on U are referred as *fuzzy open sets* in U and their complements are referred as *fuzzy closed sets* in U. The pair (U, T) is called a *fuzzy topological space (FTS)* on U.

Next it is described how the fundamental notions of *limit, continuity, compactness,* and *Hausdorff TS* can be extended to FTSs [25].

**Definition 13:** Given two FSs A and B of the FTS (U, T), B is called a *neighborhood* of A, if there exists an open FS O such that $A \subseteq O \subset B$.

**Definition 14:** A sequence $\{A_n\}$ of FSs of (U, T) converges to the FS A of (U, T), if there exists a positive integer m, such that for each integer n≥m and each neighborhood B of A we have that $A_n \subset B$. Then A is said to be the *limit* of $\{A_n\}$.

**Lemma 1:** (*Zadeh's extension principle*) Let X and Y be two non-empty crisp sets and let f: X$\rightarrow$Y be a function. Then f is extended to a function F mapping FSs in X to FSs in Y.

*Proof:* Let A be a FS in X with membership function $m_A$. Then its image F(A) is a FS B in Y with membership function $m_B$ which is defined as follows: Given y in Y, consider the set $f^{-1}(y)=\{x \in X: f(x)=y\}$. If $f^{-1}(y)=\emptyset$, then $m_B(y)=0$, and if $f^{-1}(y)\neq\emptyset$, then $m_B(y) = \max \{m_A(x): x \in f^{-1}(y)\}$. Conversely, the inverse image $F^{-1}(B)$ is the FS A in X with membership function $m_A(x)=m_B(f(x))$, for each $x \in X$.

**Definition 15:** Let (X, T) and (Y, S) be two FTSs and let f: X$\rightarrow$Y be a function. Then f is extended to a function F mapping FSs in X to FSs in Y. Then f is said to be a *fuzzily continuous* function, if, and only if, the inverse image of each open FS in Y through F is an open FS in X.

**Definition 16:** A family A=$\{A_i, i \in I\}$ of FSs of a FTS (U, T) is said to be a *cover* of U, if $U=\bigcup_{i \in I} A_i$. If the elements of A are open FSs, then A is said to be an *open cover* of U. A subset of A which is also a cover of U is called a *sub-cover* of A. The FTS (U, T) is said to be *compact*, if every open cover of U contains a sub-cover with finitely many elements.

**Definition 17:** A FTS (U, T) is said to be:

1. A *$T_1$-FTS*, if, and only if, for each pair of elements $u_1$, $u_2$ of U, $u_1 \neq u_2$, there exist at least two open FSs $O_1$ and $O_2$ such that $u_1 \in O_1$, $u_2 \notin O_1$ and $u_2 \in O_2$, $u_1 \notin O_2$.
2. A *$T_2$-FTS* (or a *separable* or a *Hausdorff FTS*), if, and only if, for each pair of elements $u_1$, $u_2$ of U, $u_1 \neq u_2$, there exist at least two open FSs $O_1$ and $O_2$ such that $u_1 \in O_1$, $u_2 \in O_2$ and $O_1 \cap O_2 = \emptyset_F$.

Obviously a $T_2$-FTS is always a $T_1$-FTS.

*6.2 Soft Topological Spaces*

Observe that the concept of FTS (Definition 12) is obtained from the classical definition of TS [45] by replacing the statement "a family of subsets of U" by the statement "a family of FSs in U". In an analogous way one can obtain the concepts of *intuitionistic FTS (IFTS)* [26], of *neutrosophic TS (NTS)* [47], of *soft TS (STS)* [48], etc. In particular, a STS is defined as follows:

**Definition 18:** A *soft topology* T on a non-empty set U is a family of SSs in U with respect to a set of parameters E such that:

- The absolute and the null soft sets $E_U$ and $E_\emptyset$ belong to T
- The intersection of any two elements of T and the union of an arbitrary number (finite or infinite) of elements of T belong also to T.

The elements of a ST T on U are said to be *open SS* and their complements are said to be *closed SS.* The triple (U, T, E) is said to be a STS on U.

Trivial examples of STs are the *discrete ST* $\{E_\emptyset, E_U\}$ and the *non-discrete ST* of all SSs in U. Reconsider also Example 1. It is straightforward to check then that T = $\{E_U, E_\emptyset, S, S_2, S_9, S_{11}\}$ is ST on U.

The concepts of limit, continuity, compactness, and Hausdorff TS are extended to STs in a way analogous to FTSs [49, 50]. In fact, Definitions 13, 14, 16 and 17 are easily turned to corresponding definitions of STSs by replacing the expression "fuzzy sets" with the expression "soft sets". For the concept of continuity we need the following Lemma ([49], definition 3.12) :

**Lemma 2:** Let (U, T. A), (V, S, B) be STSs and let u: U→V, p: A→B be given maps. Then a map $f_{pu}$ is defined with respect to u and p mapping the soft sets of T to soft sets of S.

*Proof:* If (F, A) is a soft set of T, then its image $f_{pu}((F, A))$ is a soft set of S defined by

$f_{pu}((F, A)) = (f_{pu}(F), p(A))$, where, $\forall y \in B$ is $f_{pu}(F)(y) = \bigcup_{x \in p^{-1}(y) \cap A} u(F(x))$ if $p^{-1}(y) \cap A \neq \emptyset$ and $f_{pu}(F)(y) = \emptyset$

otherwise.

Conversely, if (G, B) is a soft set of S, then its inverse image $f_{pu}^{-1}((G, B))$ is a soft set of T defined by $f_{pu}^{-1}((G, B)) = (f_{pu}^{-1}(G), p^{-1}(B))$, where $\forall x \in A$ is

$f_{pu}^{-1}(G)(x) = u^{-1}(G(p(x)))$.

**Definition 19:** Let (U, T. A), (V, S, B) be STSs and let u: U→V, p: A→B be given maps. Then the map $f_{pu}$, defined by Lemma 2, is said to be *soft pu-continuous*, if, and only if, the inverse image of each open soft set in Y through $f_{pu}$ is an open soft set in X.

**7. Discussion and Conclusions**

Three were the goals of the present review paper:

1. We came across the main steps that laid from Zadeh's FS and Atanassov's IFS to Smarandache's NS and to Molodstov's SS.
2. We presented, through suitable examples, two recently developed by us hybrid methods for assessment and DM respectively using SSs and closed real intervals (GNs) as tools.
3. We described how one can extend the concept of TS to fuzzy structures and how can define limits, continuity, compactness and Hausdorff spaces on those structures. In particular, FTSs and STSs were defined and characteristic examples were presented.

For reasons of completeness, however, we ought to note that, despite the fact that IFSs and SSs have already found many and important applications, there exist reports in the literature disputing the significance of these concepts, and in extension of the notions of IFSTS and STs, considering them as redundant, representing in an unnecessarily complicated way standard fixed-basis set theory and topology [54-57]. In the Abstract of [55], for example, one reads: "In particular, a soft set on X with a set E of parameters actually can be regarded as a 2E -fuzzy set or a crisp subset of E × X [the correct is E x P(X)]. This shows that the concept of (fuzzy) soft set is redundant". I completely disagree with this way of thinking. Adopting it, one could claim that, since a FS A in X is as subset of the Cartesian product X x m(X), where m is the membership function of A, the concept of FS is redundant!

Among probability, FSs and the other related generalizations and theories [19] there is not an ideal model for tackling effectively all the types of the existing in real world uncertainty. Each one of these theories is more suitable for dealing with special types of uncertainty, e.g. probability for randomness, FSs for vagueness, IFSs for imprecision in human thinking, NSs for ambiguity and inconsistency, etc. All these theories together, however, provide an adequate framework for managing the uncertainty.

Even more, it seems that proper combinations of the previous theories give frequently better results not only for tackling the existing uncertainty, but also for assessment purposes [42], for DM under fuzzy conditions [43] and possibly for various other human and machine activities. This is, therefore a promising area for future research.

As we have mentioned in our Introduction the concept of the ordinary FS, otherwise termed as *type-1 FS*, was generalized to the type-2 FS and further to type-n FS, n≥2, so that more uncertainty can be handled connected to the membership function [8]. The membership function of a type-2 FS is three - dimensional, its third dimension being the value of the membership function at each point of its two – dimensional domain, which is called footprint of uncertainty (FOU). The FOU is completely determined by its two bounding functions, a lower membership function and an upper membership function, both of which are type-1 FSs. When no uncertainty exists about the membership function, then a type-2 FS reduces to a type-1 FS, in a way analogous to probability reducing to determinism when unpredictability vanishes. However, when Zadeh proposed the type-2 FS in 1975 [8], the time was not right for researchers to drop what they were doing with type-1 FS and focus on type-2 FS. This changed in the late 1990s as a result of Prof. Jerry Mendel's works on type-2 FS and logic [51]. Since then, more and more researchers around the world are writing articles about type-2 FS and systems, while some important applications of type-3 FS and logic reported also recently; e.g. [52, 53]. This is, therefore another promising area for future research.

**Funding:** This research received no external funding.

**Conflicts of Interest:** The authors declare no conflict of interest.

**Acknowledgement:** The author wishes to thank the two reviewers of the paper for their useful remarks and the suggested literature.